%% file: sample.tex
\documentstyle[jair,twoside,11pt,theapa]{article}
\input epsf 
\input{psfig.sty}

\jairheading{27}{2006}{257-287}{01/06}{10/06}
\ShortHeadings{ Active Learning with Multiple Views}
{Muslea, Minton, \& Knoblock}
\firstpageno{257}

\begin{document}

\newcommand{\argmin}{\mathop{\mathrm{arg\,min}}}
\newcommand{\argmax}{\mathop{\mathrm{arg\,max}}}


\title{ Active Learning with Multiple Views}

\author{
\name Ion Muslea \email imuslea@languageweaver.com  \\
\addr Language Weaver, Inc.  \\
      4640 Admiralty Way, Suite 1210      \\
      Marina del Rey, CA 90292
\AND
\name Steven Minton \email minton@fetch.com \\
\addr Fetch Technologies, Inc. \\
      2041 Rosecrans Ave., Suite 245 \\
      El Segundo, CA 90245
\AND
\name Craig A. Knoblock \email knoblock@isi.edu \\
\addr University of Southern California \\
      4676 Admiralty Way \\
      Marina del Rey, CA 90292 
}


\maketitle

\begin{abstract}


{\em Active learners} alleviate the burden of labeling large amounts
of data by detecting and asking the user to label only the most
informative examples in the domain.  We focus here on active learning
for {\em multi-view} domains, in which there are several disjoint
subsets of features ({\em views}), each of which is sufficient to
learn the target concept.  In this paper we make several
contributions. First, we introduce Co-Testing, which is the first
approach to multi-view active learning.  Second, we extend the
multi-view learning framework by also exploiting {\em weak views},
which are adequate only for learning a concept that is more
general/specific than the target concept.  Finally, we empirically
show that Co-Testing outperforms existing active learners on a variety
of real world domains such as wrapper induction, Web page
classification, advertisement removal, and discourse tree parsing.


\end{abstract}

\input{intro}

\input{background}

\input{related}

\input{coTst}

\input{cmpRelWork}

\input{apps}

\input{Conclusions}

\section* {Acknowledgments}

This research is based upon work supported in part by the National Science
Foundation under Award No. IIS-0324955 and grant number 0090978, in part by
the Defense Advanced Research Projects Agency (DARPA), through the
Department of the Interior, NBC, Acquisition Services Division, under
Contract No. NBCHD030010, and in part by the Air Force Office of Scientific
Research under grant number FA9550-04-1-0105.  The U.S.Government is
authorized to reproduce and distribute reports for Governmental purposes
notwithstanding any copy right annotation thereon.  The views and
conclusions contained herein are those of the authors and should not be
interpreted as necessarily representing the official policies or
endorsements, either expressed or implied, of any of the above organizations
or any person connected with them.



\vskip 0.2in
\bibliography{activeLearning,wg,ie,Clumpiness,vValid,thesis,RW}
\bibliographystyle{theapa}

\end{document}

%% file: intro.tex
\section{ Introduction }

Labeling the training data for a machine learning algorithm is a tedious,
time consuming, error prone process; furthermore, in some application
domains, the labeling of each example may also be extremely expensive (e.g.,
it may require running costly laboratory tests).  {\em Active learning}
algorithms \cite{calSelSamp,roy01toward,tong01Support} cope with this
problem by detecting and asking the user to label only the most informative
examples in the domain, thus reducing the user's involvement in the data
labeling process.

In this paper, we introduce Co-Testing, an active learning technique
for {\em multi-view} learning tasks; i.e., tasks that have several
disjoint subsets of features ({\em views}), each of which is
sufficient to learn the concepts of interest. For instance, Web page
classification is a multi-view task because Web pages can be
classified based on the words that appear {\em either} in the
documents {\em or} in the hyperlinks pointing to them \cite{coTra};
similarly, one can classify segments of televised broadcast based {\em
  either} on the video {\em or} on the audio information, or one can
perform speech recognition based on {\em either} sound {\em or} lip
motion features \cite{speech-2V}.

Co-Testing is a two-step iterative algorithm that requires as input a few
labeled and many unlabeled examples. First, Co-Testing uses the few labeled
examples to learn a hypothesis in each view. Then it applies the learned
hypotheses to all unlabeled examples and detects the set of {\em contention
points} (i.e., unlabeled examples on which the views predict a different
label); finally, it {\em queries} (i.e., asks the user to label) one of the
contention points, adds the newly labeled example to the training set, and
repeats the whole process.  Intuitively, Co-Testing relies on the following
observation: if, for an unlabeled example, the hypotheses learned in each
view predict a different label, at least one of them makes a mistake on that
particular prediction. By asking the user to label such a contention point,
Co-Testing is guaranteed to provide useful information for the view that
made the mistake.


In this paper we make several contributions. First, we introduce
Co-Testing, a family of active learners for multi-view learning
tasks. Second, we extend the traditional multi-view learning framework
by also allowing the use of {\em weak views}, in which one can
adequately learn only a concept that is strictly more general or more
specific than the target concept (all previous multi-view work makes
the {\em strong view} assumption that each view is adequate for
learning the target concept).  Last but not least, we show that, in
practice, Co-Testing clearly outperforms existing active learners on a
variety of real world domains such as wrapper induction, Web page
classification, advertisement removal, and discourse tree parsing.

Compared with previous work, Co-Testing is unique in several ways:
\begin{description}
\item[1.] existing multi-view approaches \cite{coTra,coBoost,pierce01Limit},
which also use a small set of labeled and a large set of unlabeled examples,
are based on the idea of bootstrapping the views from each other. In
contrast, Co-Testing is the first algorithm that exploits multiple views for
active learning purposes. Furthermore, Co-Testing allows the simultaneous
use of strong and weak views without additional data engineering costs.
\item[2.] existing active learners, which pool all domain features together,
are typically designed to exploit some properties specific to a particular
(type of) {\em base learner} (i.e., the algorithm used to learn the target
concept); for example, {\em uncertainty reduction} methods assume that the
base learner provides a reliable estimate of its confidence in each
prediction. In contrast, Co-Testing uses the multiple views to detect the
contention points, among which it chooses the next query. This approach has
several advantages:
\begin{description}
\item[-] it converges quickly to the target concept because it is based on
the idea of {\em learning from mistakes} (remember that each contention
point is guaranteed to represent a mistake in at least one of the
views). In contrast, existing active learners often times query examples
that are classified correctly, but with a low confidence.
\item[-] in its simplest form (i.e., Naive Co-Testing, which is described
in section \ref{coTest}), it makes {\em no assumptions} about the properties
of the base learner. More precisely, by simply querying an arbitrary
contention point, Co-Testing is guaranteed to provide ``the mistaken view''
with a highly informative example.
\item[-] by considering only the contention points as query candidates, it
allows the use of query selection heuristics that - computationally - are
too expensive to be applied to the entire set of unlabeled examples.
\end{description}
\end{description}

The remainder of the paper is organized as follows. First, we
introduce the concepts and notation, followed by a comprehensive
survey of the literature on active and multi-view learning. Then we
formally introduce the Co-Testing family of algorithms and we present
our empirical evaluation on a variety of real-world domains.


%% file: background.tex
\section{ Preliminaries: Terminology and Notation }

For any given learning task, the set of all possible domain examples is
called the {\em instance space} and is denoted by $X$.  Any $x \in X$
represents a particular {\em example} or {\em instance}. In this paper we
are concerned mostly with examples that are represented as {\em feature
vectors} that store the values of the various {\em attributes} or {\em
features} that describe the example.

The concept to be learned is called the {\em target concept}, and it
can be seen as a function $c: X \rightarrow \{l_1,l_2,\ldots,l_N\}$
that classifies any instance $x$ as a member of one of the $N$ classes
of interest $l_1,l_2,\ldots,l_N$. In order to learn the target
concept, the user provides a set of {\em training examples}, each of
which consists of an instance $x \in X$ and its label, $c(x)$. The
notation $\langle x, c(x) \rangle$ denotes such a training
example. The symbol $L$ is used to denote the set of labeled training
examples (also known as the {\em training set}).

Given a training set $L$ for the target concept $c$, an inductive
learning algorithm ${\cal L}$ searches for a function $h: X
\rightarrow \{l_1,l_2,\ldots,l_N\}$ such that $\forall x \in X, h(x) =
c(x)$. The learner ${\cal L}$ searches for $h$ within the set $H$ of
all possible hypotheses, which is (typically) determined by the person
who designs the learning algorithm. A hypothesis $h$ is {\em
consistent} with the training set $L$ if and only if $\forall \langle
x, c(x)\rangle \in L, h(x) = c(x)$. Finally, the {\em version space}
$VS_{H,L}$ represents the subset of hypotheses in $H$ that are
consistent with the training set $L$.

By definition, a {\em passive learning} algorithm takes as input a randomly
chosen training set $L$. In contrast, {\em active learning} algorithms have
the ability to choose the examples in $L$. That is, they detect the most
informative examples in the instance space $X$ and ask the user to label
only them; the examples that are chosen for labeling are called {\em
queries}. In this paper we focus on {\em selective sampling} algorithms,
which are active learners that choose the queries from a given {\em working
set} of unlabeled examples $U$ (we use the notation $\langle x, ? \rangle$
to denote an unlabeled examples). In this paper the terms {\em active
learning} and {\em selective sampling} are used interchangeably.

In the traditional, {\em single-view} machine learning scenario, a learner
has access to the entire set of domain features. By contrast, in the {\em
multi-view} setting one can partition the domain's features in subsets ({\em
views}) that are {\em sufficient} for learning the target concept.  Existing
multi-view learners are {\em semi-supervised} algorithms: they exploit
unlabeled examples to boost the accuracy of the classifiers learned in each
view by bootstrapping the views from each other.

In multi-view learning, an example $x$ is described by a different set of
features in each view. For example, in a domain with $k$ views $V_1, V_2,
\ldots V_k$, a labeled example can be seen as a tuple $\langle x_1, x_2,
\ldots, x_k, l\rangle$, where $l$ is its label, and $x_1, x_2, \ldots, x_k$
are its descriptions in the $k$ views. Similarly, a $k$-view unlabeled
example is denoted by $\langle x_1, x_2, \ldots, x_k, ? \rangle$. For any
example $x$, $V_i(x)$ denotes the descriptions $x_i$ of $x$ in $V_i$.
Similarly, $V_i(L)$ consists of the descriptions in $V_i$ of all the
examples in $L$.

%% file: related.tex
\section{ Background on Active and Multi-view Learning}

Active learning can be seen as a natural development from the earlier work
on optimum experimental design \cite{fedorov72TOE}.  In the early 1980s, the
machine learning community started recognizing the advantages of inductive
systems that are capable of querying their instructors. For example, in
order to detect errors in Prolog programs, the Algorithmic Debugging System
\cite{shapiro81General,shapiro82APDiag} was allowed to ask the user several
types of queries. Similarly, concept learning systems such as Marvin
\cite{sammut86LCbAQ} and {\sc cat} \cite{gross91CAT} used queries as an
integral part of their respective learning strategies.

Our literature review below is structured as follows. First, we discuss the
early, mostly theoretical results on {\em query construction}. Then we focus
on selective sampling algorithms, which select as the next query one of the
unlabeled examples from the working set. Finally, we conclude by reviewing
the existing multi-view learning algorithms.

\subsection{ Active Learning by Query Construction}

The earliest approaches to formalizing active learning appeared in the
seminal papers of \citeauthor{angluin82ANote}
\citeyear{angluin82ANote,angluin88QCL} and \citeauthor{valiant84ToL}
\citeyear{valiant84ToL}, who focused on exact concept induction and
learning in the {\sc pac} framework, respectively. This theoretic work
focused on learning classes of concepts such as regular sets, monotone
{\sc dnf} expressions, and $\mu-$expressions. Besides {\em membership
queries} such as ``{\small is this an example of the target
concept?},'' Angluin also used more sophisticated types of queries
such as {\em equivalence queries} (``{\small is this concept
equivalent with the target concept?}'') or {\em superset queries}
(``{\small is this concept a superset of the target concept?}'').

These early active learners took a {\em constructive} approach to
query generation in the sense that each query is ({\em artificially})
constructed by setting the values of the attributes so that the query
is as informative as possible.  In practice, this may raise some
serious problems; for example, consider a hand-writing recognizer that
must discriminate between the 10 digits \cite{lang92Query}. In this
scenario, an informative query may consist of an image that represents
a ``fusion'' of two similarly-looking digits, such as ``3'' and ``5.''
When presented with such an image, a user cannot label it properly
because it does not represent a recognizable digit. Consequently, a
query is ``wasted'' on a totally irrelevant image. Similar situations
appear in many real world tasks such as text classification,
information extraction, or speech recognition: whenever the active
learner artificially builds a query for such a domain, it is highly
unlikely that the newly created object has any meaning for the human
user.

Despite this practical applicability issue, the constructive approach
to active learning leads to interesting theoretical insights about the
merits of various types of queries. For example, researchers
considered learning with:
\begin{description}
\item[-] {\em incomplete queries}, for which the query's answer may be
``{\tt I don't know.}''
\cite{angluin91randomly,goldman92learning,sloan94learning,blum98learning};
\item[-] {\em malicious queries}, for which the answer to the queries
may be erroneous \cite{angluin97malicious,angluin94malicious,angluin94exact}.
\end{description}

New learning problems were also considered, from unrestricted {\sc
dnf} expressions \cite{jackson94efficient,blum94weakly} and unions of
boxes \cite{goldberg94learning} to tree patterns
\cite{amoth98exact,amoth99exact} and Horn clauses
\cite{reddy97learning}. Researchers also reported results on 
applying active learning to neural networks
\cite{hwang91QBased,baum91NNA,watkin92SEfP,hasenj98active}
and for combining declarative bias (prior knowledge) and active
learning \cite{tadepalli93learning,tadepalli98Learning}.

\subsection{ Selective Sampling}
\label{ssRW}

{\em Selective sampling} represents an alternative active learning
approach. It typically applies to {\em classification} tasks in which
the learner has access to a large number of unlabeled examples. In
this scenario, rather than constructing an informative query, the
active learner asks the user to label one of the existing unlabeled
examples. Depending on the {\em source} of unlabeled examples, there
are two main types of sampling algorithms: stream- and pool- based.
The former assumes that the active learner has access to an (infinite)
stream of unlabeled examples \cite{FSST,cbProbabClas,ssNL3}; as
successive examples are presented to it, the active learner must
decide which of them should be labeled by the user. In contrast, in
the pool-based scenario \cite{LewGal,LewCat,emAL,coTst,coEMT}, the
learner is presented with a {\em working set} of unlabeled examples;
in order to make a query, the active learner goes through the entire
pool and selects the example to be labeled next.

Based on the criterion used to select the next query, selective
sampling algorithms fall under three main categories:
\begin{description}

\item[-] {\em uncertainty reduction}: the system queries the
example on which the current hypothesis makes the least confident
prediction;

\item[-] {\em expected-error minimization}: the system queries the
example that maximizes the {\em expected} reduction in classification
error;

\item[-] {\em version space reduction}: the system queries the example
that, once labeled, removes as much as possible of the version space.
\end{description}

The {\em uncertainty reduction} approach to selective sampling works as
follows: first, one uses the labeled examples to learn a classifier; then
the system queries the unlabeled example on which this classifier makes the
{\em least confident} prediction. This straightforward idea can be applied
to any base learner for which one can reliably estimate the confidence of
its predictions. Confidence-estimation heuristics were proposed for a
variety of base learners such as logistic regression \cite{LewGal,LewCat},
partially hidden Markov Models \cite{scheffer01active}, support vector
machines \cite{schohn00less,campbell00query}, and inductive logic
programming \cite{alieICML99}.

The second, more sophisticated approach to selective sampling, {\em
  expected-error minimization}, is based on the {\em statistically
  optimal} solution to the active learning problem.  In this scenario,
the intuition is to query the unlabeled example that minimizes the
error rate of the (future) classifier on the test set. Even though for
some (extremely simple) base learners one can find such optimal
queries \cite{cohn96active}, this is not true for most inductive
learners. Consequently, researchers proposed methods to {\em estimate}
the error reduction for various types of base learners. For example,
\citeauthor{roy01toward} \citeyear{roy01toward} use a sample
estimation method for the Naive Bayes classifier; similar approaches
were also described for parameter learning in Bayesian nets
\cite{tong00active} and for nearest neighbor classifiers
\cite{lindenbaum04selective}.

The heuristic approach to {\em expected-error minimization} can be
summarized as follows. First, one chooses a {\em loss function}
\cite{roy01toward} that is used to estimate the future error rate. Then each
unlabeled example $x$ in the working set is considered as the possible next
query, and the system estimates the expected reduction of the error rate for
each possible label that $x$ may take. Finally, the system queries the
unlabeled example that leads to the largest estimated reduction in the error
rate.

Finally, a typical {\em version space reduction} active learner works as
follows: it generates a {\em committee} of several hypotheses, and it
queries the unlabeled examples on which the disagreement within the
committee is the greatest. In a two-class learning problem, this strategy
translates into making queries that remove approximately half of the version
space. Depending on the method used to generate the committee, one can
distinguish several types of active learners:
\begin{description}

\item[-] Query-by-Committee selects a committee by randomly sampling
hypotheses from the version space. Query-by-Committee was applied to a
variety of base learners such as perceptrons \cite{FSST}, Naive Bayes
\cite{emAL}, and Winnow \cite{alcTC}. Furthermore, Argamon-Engelson
and Dagan \citeyear{cbProbabClas,ssNL3} introduce an extension to
Query-by-Committee for Bayesian learning. In the Bayesian framework,
one can create the committee by sampling classifiers according to
their posterior distributions; that is, the better a hypothesis
explains the training data, the more likely it is to be sampled.  The
main limitation of Query-by-Committee is that it can be applied only
to base learners for which it is feasible to randomly sample
hypotheses from the version space.

\item[-] {\sc sg}-net \cite{calSelSamp} creates a 2-hypothesis
committee that consists of a ``{\em most-general}'' and a ``{\em
most-specific}'' classifier. These two hypotheses are generated by
{\em modifying the base learner} so that it learns a classifier that
labels as many as possible of the unlabeled examples in the working
set as positive or negative, respectively. This approach has an
obvious drawback: it requires the user to modify the base learner so
that it can generate ``{\em most-general}'' and ``{\em
most-specific}'' classifiers.

\item[-] Active-Decorate \cite{prem04al} can be seen as both a
generalization and an improvement of {\sc sg}-net. It generates a
large and diverse committee by successively augmenting the original
training set with additional sets of artificially-generated
examples. More precisely, it generates artificial examples in keeping
with the distribution of the instance space; then it applies the
current committee to each such example, and it labels the artificial
example with the label that contradicts most of the committee's
predictions. A new classifier is learned from this augmented dataset,
and then the entire process is repeated until the desired committee
size is reached. Active-Decorate was successfully used for domains
with nominal and numeric features, but it is unclear how it could be
applied to domains such as text classification or extraction, where
generating the artificial examples may be problematic.

\item[-] Query-by-Bagging and Query-by-Boosting \cite{qBB} create the
committee by using the well-known bagging \cite{bagging} and boosting
\cite{boosting} algorithms, respectively. These algorithms were
introduced for the {\sc c4.5} base learner, for which both bagging and
boosting are known to work extremely well.

\end{description}

In general, committee-based sampling tends to be associated with the
{\em version space reduction} approach. However, for base learners
such as support vector machines, one can use a {\em single hypothesis}
to make queries that remove (approximately) half of the version space
\cite{tong01Support}. Conversely, committee-based sampling can also be
seen as relying on the {\em uncertainty reduction} principle: after
all, the unlabeled example on which the disagreement within the
committee is the greatest can be also seen as the example that has the
least certain classification.

\subsection{ Multi-view, Semi-supervised Learning}
\label{coTraRW}

As already mentioned, Blum and Mitchell \citeyear{coTra} provided the first
formalization of learning in the multi-view framework.  Previously, this
topic was largely ignored, though the idea clearly shows up in applications
such as word sense disambiguation \cite{yarowsky-2v} and speech recognition
~\cite{speech-2V}.  Blum and Mitchell proved that two independent,
compatible views can be used to {\sc pac}-learn \cite{valiant84ToL} a
concept based on few labeled and many unlabeled examples. They also
introduced Co-Training, which is the first general-purpose, multi-view
algorithm.

Collins and Singer \citeyear{coBoost} proposed a version of
Co-Training that is biased towards learning hypotheses that predict
the same label on most of the unlabeled examples. They introduce an
explicit objective function that measures the compatibility of the
learned hypotheses and use a boosting algorithm to optimize this
objective function.  In a related paper \cite{dasgupta01}, the authors
provide {\sc pac}-like guarantees for this novel Co-Training algorithm
(the assumption is, again, that the views are both independent and
compatible).  Intuitively, \citeauthor{dasgupta01}
\citeyear{dasgupta01} show that the ratio of contention points to
unlabeled examples is an upper-bound on the error rate of the
classifiers learned in the two views.

Abney \citeyear{abney02} extends the work of \citeauthor{dasgupta01} by
relaxing the view independence assumption.  More precisely, the author shows
that even with views that are {\em weakly dependent}, the ratio of
contention points to unlabeled examples still represents an upper-bound on
the two view's error rate. Unfortunately, this paper introduces just a
theoretical definition for the {\em weak dependence} of the views, without
providing an intuitive explanation of its practical consequences.

Researchers proposed two main types of extensions to the original
Co-Training algorithm: modifications of the actual algorithm and
changes aiming to extend its practical applicability. The former cover
a wide variety of scenarios:
\begin{description}

\item[-] Co-EM \cite{coEM,coEMSVM} uses Expectation Maximization
  \cite{em} for multi-view learning. Co-EM can be seen as the closest
  implementation of the theoretical framework proposed by
  \citeauthor{coTra} \citeyear{coTra}.

\item[-] Ghani \citeyear{coGhaniICML02} uses Error-Correcting Output Codes
to allow Co-Training and Co-EM to scale up to problems with a large number
of classes.

\item[-] Corrected Co-Training \cite{pierce01Limit} asks the user to
manually correct the labels of the bootstrapped examples. This approach is
motivated by the observation that the quality of the bootstrapped data is
crucial for Co-Training's convergence.

\item[-] Co-Boost \cite{coBoost} and Greedy Agreement \cite{abney02} are
Co-Training algorithms that explicitly aim to minimize the number of
contention points.

\end{description}

The second group of extensions to Co-Training is motivated by the fact
that, in practice, one also encounters many problems for which there
is no straightforward way to split the features in two views.  In
order to cope with this problem, \citeauthor{goldman00enhancing}
\citeyear{goldman00enhancing} advocate the use of {\em multiple
biases} instead of multiple views. The authors introduce an algorithm
similar to Co-Training, which bootstraps from each other hypotheses
learned by two different base learners; this approach relies on the
assumption that the base learners generate hypotheses that partition
the instance space into equivalence classes. In a recent paper,
\citeauthor{zhou04ictai} ~\citeyear{zhou04ictai} use the idea of a
multi-biased committee for active learning; i.e., they use various
types base learners to obtain a diverse committee, and then query the
examples on which this committee disagree the most.

Within the multi-view framework, Nigam and Ghani \citeyear{coEM} show
that, for ``bag-of-words'' text classification, one can create two
views by arbitrarily splitting the original set of features into two
sub-sets.  Such an approach fits well the text classification domain,
in which the features are abundant, but it is unlikely to work on
other types of problems.  An alternative solution is proposed by
\citeauthor{raskutti02Comb} \citeyear{raskutti02Comb}, where the
authors create a second view that consists of a variety of features
that measure the examples' similarity with the $N$ largest clusters in
the domain. Finally, \citeauthor{vValidICML02} \citeyear{vValidICML02}
propose a meta-learning approach that uses past experiences to predict
whether the given views are appropriate for a new, unseen learning
task.

%% file: coTst.tex
\section{ The Co-Testing Family of Algorithms }
\label{coTest}

In this section, we discuss in detail the Co-Testing family of
algorithms.  As we already mentioned, Co-Testing can be seen as a
two-step iterative process: first, it uses a few labeled examples to
learn a hypothesis in each view; then it queries an unlabeled example
for which the views predict different labels. After adding the queried
example to the training set, the entire process is repeated for a
number of iterations.

The remainder of this section is organized as follows: first, we
formally present the Co-Testing family of algorithms and we discuss
several of its members. Then we introduce the concepts of strong and
weak views, and we analyze how Co-Testing can exploit both types of
views (previous multi-view learners could only use strong
views). Finally, we compare and contrast Co-Testing to the related
approaches.

\begin{table}[thp]
\caption[The Co-Testing family of algorithms]{ The Co-Testing family
of algorithms: repeatedly learn a classifier in each view and query an
example on which they predict different labels.}  
\hrule
\begin{tabbing}
 {\bf Gi}\={\bf ve}\={\bf n:} \\ 
 \>- a base learner ${\cal L}$ \\
 \>- a learning domain with features $V = \{a_1, a_2, \ldots, a_N\}$ \\
 \>- $k$ views $V_1, V_2, \ldots, V_k$ such that \(V = {\displaystyle \bigcup_{i=1}^{k}} V_i\) and
 $\forall i,j \in \{1, 2, \ldots, k\}, i\neq j, V_i \cap V_j = \O$ \\
 \>- the sets $L$ and $U$ of labeled and unlabeled examples, respectively \\
 \>- number $N$ of queries to be made \\ 
\\
- {\tt L}\={\tt O}\={\tt O}\={\tt P} for $N$ iterations \\
\>\>\>- use ${\cal L}$ to learn the classifiers $h_1, h_2, \ldots, h_k $ in the views $V_1, V_2, \ldots, V_k$, respectively \\
\>\>\>- let $ContentionPoints$ = \{ $\langle x_1, x_2, \ldots, x_k, ? \rangle \in U \mid \exists i, j$ $h_i(x_i) \neq h_j(x_j)$ \}\\
\>\>\>- let $\langle x_1, x_2, \ldots, x_k, ? \rangle$ = {\bf SelectQuery}($ContentionPoints$)\\
\>\>\>- remove $\langle x_1, x_2, \ldots, x_k, ? \rangle$ from  $U$ and ask for its label $l$ \\
\>\>\>- add $\langle x_1, x_2, \ldots, x_k, l \rangle$ to $L$ \\
\\
- $h_{OUT}$ = {\bf CreateOutputHypothesis}( $h_1, h_2, \ldots, h_k $ )
\end{tabbing}
\label{coTST}
\end{table}

\subsection{ The Family of Algorithms }

Table \ref{coTST} provides a formal description on the Co-Testing family of
algorithms. The input consists of $k$ views $V_1, V_2, \ldots, V_k$, a base
learner ${\cal L}$, and the sets $L$ and $U$ of labeled and unlabeled
examples, respectively. Co-Testing algorithms work as follows: first, they
learn the classifiers $h_1, h_2, \ldots, h_k $ by applying the algorithm
${\cal L}$ to the projection of the examples in $L$ onto each view. Then
they apply $h_1, h_2, \ldots, h_k$ to all unlabeled examples in $U$ and
create the set of contention points, which consists of all unlabeled
examples for which at least two of these hypotheses predict a different
label. Finally, they query one of the contention points and then repeat the
whole process for a number of iterations. After making the allowed number of
queries, Co-Testing creates an {\em output hypothesis} that is used to make
the actual predictions.

The various members of the Co-Testing family differ from each other
with two respects: the strategy used to select the next query, and the
manner in which the output hypothesis is constructed. In other words,
each Co-Testing algorithm is uniquely defined by the choice of the
functions {\bf SelectQuery()} and {\bf CreateOutputHypothesis()}.

In this paper we consider three types of query selection
strategies:
\begin{description}
\item[-] {\em naive}: choose at random one of the contention points.
This straightforward strategy is appropriate for base learners that
lack the capability of reliably estimating the confidence of their
predictions. As this naive query selection strategy is independent of
both the domain and the base learner properties, it follows that it
can be used for solving any multi-view learning task.

\item[-] {\em aggressive}: choose as query the contention point $Q$ on
which the least confident of the hypotheses $h_1, h_2, \ldots, h_k$
makes the most confident prediction; more formally,

\[Q = \argmax_{x \in ContentionPoints} \;\; \min_{i \in \{1, 2, \ldots, k\} } {\small Confidence}(h_i(x))\]

{\em Aggressive Co-Testing} is designed for high
accuracy domains, in which there is little or no noise. On such domains,
discovering unlabeled examples that are misclassified ``with high
confidence'' translates into queries that remove significantly more than
half of the version space.

\item[-] {\em conservative}: choose the contention point on which the
confidence of the predictions made by $h_1, h_2, \ldots, h_k$ is as
close as possible (ideally, they would be equally confident in
predicting different labels); that is,

\[Q = \argmin_{x \in ContentionPoints} \left(
\max_{f \in \{h_1, \ldots , h_k\}} {\small Confidence}(f(x)) \; - 
\min_{g \in \{h_1, \ldots , h_k\}}
{\small Confidence}(g(x)) \right) \]  

{\em Conservative Co-Testing} is appropriate for noisy domains, where
the {\em aggressive} strategy may end up querying mostly noisy
examples.

\end{description}

Creating the output hypothesis also allows the user to choose from a
variety of alternatives, such as:
\begin{description}

\item[-] {\em weighted vote}: combines the vote of each hypothesis,
weighted by the confidence of their respective predictions.

\[h_{OUT}(x) = \argmax_{l \in Labels} \;\;
\sum_{{\scriptsize
\begin{array}{c}
g \in \{h_1, \ldots , h_k\} \\
{g(x) = l}
\end{array}}} 
{\small Confidence}(g(x)) \]

\item[-] {\em majority vote}: Co-Testing chooses the label that was
predicted by most of the hypotheses learned in the $k$ views.

\[h_{OUT}(x) = \argmax_{l \in Labels} \;\;
\sum_{{\scriptsize
\begin{array}{c}
g \in \{h_1, \ldots , h_k\} \\
{g(x) = l}
\end{array}}} 
1 \]  

This strategy is appropriate when there are at least three views, and
the base learner cannot reliably estimate the confidence of its
predictions.

\item[-] {\em winner-takes-all}: the output hypothesis is the one
learned in the view that makes the smallest number of mistakes over
the $N$ queries. This is the most obvious solution for 2-view learning
tasks in which the base learner cannot (reliably) estimate the
confidence of its predictions. If we denote by {\small $Mistakes(h_1),
Mistakes(h_2), \ldots , Mistakes(h_k)$} the number of mistakes made by
the hypotheses learned in the $k$ views on the $N$ queries, then

\[h_{OUT}(x) = \argmin_{g \in \{h_1, \ldots , h_k\}} {\small Mistakes(g)}\]

\end{description}

\subsection{ Learning with Strong and Weak Views}

In the original multi-view setting \cite{coTra,coTst}, one makes the {\em
strong views} assumption that each view is sufficient to learn the target
concept.  However, in practice, one also encounters views in which one can
accurately learn only a concept that is strictly {\em more general} or {\em
more specific} than the concept of interest \cite{muslea2003}. This is often
the case in domains that involve {\em hierarchical} classification, such as
information extraction or email classification. For example, it may be
extremely easy to discriminate (with a high accuracy) between work and
personal emails based solely on the email's sender; however, this same
information may be insufficient for predicting the work or personal {\em
sub-folder} in which the email should be stored.

We introduce now the notion of a {\em weak view}, in which one can
accurately learn only a concept that is strictly {\em more general} or {\em
more specific} than the target concept. Note that learning in a weak view is
qualitatively different from learning ``an approximation'' of the target
concept: the latter represents learning with imperfect features, while the
former typically refers to a (easily) learnable concept that is a strict
generalization/specialization of the target concept (note that, in the real
world, imperfect features and noisy labels affect learning in both strong
and weak views).

In the context of learning with strong and weak views, we redefine {\em
contention points} as the unlabeled examples on which the {\em strong views}
predict a different label. This is a necessary step because of two reasons:
first, as the weak view is inadequate for learning the target concept, it
typically disagrees with the strong views on a large number of unlabeled
examples; in turn, this would increase the number of contention points and
skew their distribution. Second, we are not interested in fixing the
mistakes made by a weak view, but rather in using this view as an additional
information source that allows faster learning in the strong views (i.e.,
from fewer examples).

Even though the weak views are inadequate for learning the target concept,
they can be exploited by Co-Testing both in the {\bf SelectQuery()} and {\bf
CreateOutputHypothesis()} functions. In particular, weak views are extremely
useful for domains that have only two strong views: 
\begin{description}
\item[-] the weak view can be used in {\bf CreateOutputHypothesis()} as a
{\em tie-breaker} when the two strong views predict a different label.

\item[-] {\bf SelectQuery()} can be designed so that, ideally, each query
would represent a mistake in both strong views. This can be done by first
detecting the contention points - if any - on which the weak view disagrees
with both strong views; among these, the next query is the one on which the
weak view makes the most confident prediction.  Such queries are likely to
represent a mistake in {\em both} strong views, rather than in just one of
them; in turn, this implies simultaneous large cuts in {\em both} strong
version spaces, thus leading to faster convergence.
\end{description}

In section \ref{AL4WI} we describe a Co-Testing algorithm that exploits
strong and weak views for wrapper induction domains
\cite{muslea2003,stalkerJAAMAS}.  Note that learning from strong and weak
views clearly extends beyond wrapper induction tasks: for example, the idea
of exploiting complementary information sources (i.e., different types of
features) appears in the two multi-strategy learners
\cite{nick01IEbyTC,nahm01dicotex} that are discussed in section
\ref{wvRelW}.

%% file: cmpRelWork.tex
\subsection{ Co-Testing {\em vs.} Related Approaches  }

As we already mentioned in section \ref{coTraRW}, existing multi-view
approaches are typically semi-supervised learners that bootstrap the views
from each other. The two exceptions \cite{coEMT,jones03al4ie} interleave
Co-Testing and Co-EM \cite{coEM}, thus combining the best of both worlds:
semi-supervised learning provides the active learner with more accurate
hypotheses, which lead to more informative queries; active learning provides
the semi-supervised learner with a more informative training set, thus
leading to faster convergence.

\subsubsection{ Co-Testing {\em vs.} Existing Active Learners}
\label{coTstVsAL}

Intuitively, Co-Testing can be seen as a committee-based active learner that
generates a committee that consists of one hypothesis in each view. Also
note that Co-Testing can be combined with virtually any of the existing
single-view active learners: among the contention points, Co-Testing can
select the next query based on any of the heuristics discussed in section
\ref{ssRW}.

There are two main differences between Co-Testing and other active learners:
\begin{description}
\item[-] except for Co-Testing and its variants
~\shortcite{coEMT,jones03al4ie}, all other active learners work in the
single-view framework (i.e., they pool together all the domain features).
\item[-] single-view active learners are typically designed for a particular
(class of) base learner(s). For example, Query-by-Committee \cite{sosQBC}
assumes that one can randomly sample hypotheses from the version space,
while Uncertainty Sampling \cite{LewGal,LewCat} relies on the base learner's
ability to reliably evaluate the confidence of its predictions. In contrast,
the basic idea of Co-Testing (i.e., querying contention points) applies to
any multi-view problem, independently of the base learner to be used.
\end{description}

The Co-Testing approach to active learning has both advantages and
disadvantages. On one hand, Co-Testing cannot be applied to problems that do
not have at least two views. On the other hand, for any multi-view problem,
Co-Testing can be used with the best base learner for that particular
task. In contrast, in the single-view framework, one often must either
create a new active learning method for a particular base learner or, even
worse, modify an existing base learner so that it can be used in conjunction
with an existing sampling algorithm.

To illustrate this last point, let us briefly consider learning for {\em
information extraction}, where the goal is to use machine learning for
extracting relevant strings from a collection of documents (e.g., extract
the perpetrators, weapons, and victims from a corpus of news stories on
terrorist attacks). As information extraction is different in nature from a
typical classification task, existing active learners cannot be applied in a
straightforward manner:
\begin{description}
\item[-] for {\em information extraction from free text} ({\sc ie}),
  the existing active learners
  \cite{alieICML99,WHISK,scheffer01active} are crafted based on
  heuristics specific to their respective base learners, {\sc rapier},
  {\sc whisk}, and Partially Hidden Markov Models. An alternative is
  discussed by ~\citeauthor{kushmerick03al4ie}
  ~\citeyear{kushmerick03al4ie}, who explore a variety of {\em {\sc
      ie}-specific} heuristics that can be used for active learning
  purposes and analyze the trade-offs related to using these
  heuristics.
\item[-] for {\em wrapper induction}, where the goal is to extract data from
Web pages that share the same underlying structure, there are no reported
results for applying (single-view) active learning. This is because typical
wrapper induction algorithms
~\cite{stalkerJAAMAS,KushmerickAIJ,SoftMealyJIS98} are base learners that
lack the properties exploited by the single-view active learners reviewed in
section \ref{ssRW}.: they are {\em determinist} learners that are {\em noise
sensitive}, provide {\em no confidence} in their predictions, and make {\em
no mistakes} on the training set.
\end{description}
\noindent In contrast, Co-Testing applies naturally to both wrapper
induction \shortcite{coTst,muslea2003} and information extraction from free
text \shortcite{jones03al4ie}. This is due to the fact that Co-Testing does
{\em not} rely on the base learner's properties to identify its highly
informative set of candidate queries; instead, it focuses on the contention
points, which, by definition, are {\em guaranteed} to represent mistakes in
some of the views.

\subsubsection{ Exploiting Weak Views}
\label{wvRelW}

We briefly discuss now two learning tasks that can be seen as learning from
strong and weak views, even though they were not formalized as such, and the
views were not used for active learning. An additional application domain
with strong and weak views, wrapper induction, is discussed at length in
section \ref{experimWI}.


The {\sc discotex} \cite{nahm01dicotex} system was designed to extract
job titles, salaries, locations, etc from computer science job
postings to the newsgroup {\small {\tt austin.jobs}}. {\sc discotex}
proceeds in four steps: first, it uses {\sc rapier}
\cite{rapierAIII99} to learn extraction rules for each item of
interest. Second, it applies the learned rules to a large, unlabeled
corpus of job postings and creates a database that is populated with
the extracted data. Third, by text mining this database, {\sc
discotex} learns to predict the value of each item based on the values
of the other fields; e.g., it may discover that ``{\small {\tt IF the
job requires {\sc c++} and {\sc corba} THEN the development platforms
include Windows.}}'' Finally, when the system is deployed and the {\sc
rapier} rules fail to extract an item, the mined rules are used to
predict the item's content.

In this scenario, the {\sc rapier} rules represent the {\em strong
  view} because they are sufficient for extracting the data of
interest. In contrast, the mined rules represent the {\em weak view}
because they cannot be learned or used by themselves. Furthermore, as
{\sc discotex} discards all but the most accurate of the mined rules,
which are highly-specific, it follows that this weak view is used to
learn concepts that are {\em more specific} than the target concept.
\citeauthor{nahm01dicotex} \citeyear{nahm01dicotex} show that the
mined rules improve the extraction accuracy by capturing information
that complements the {\sc rapier} extraction rules.

Another domain with strong and weak views is presented by
\citeauthor{nick01IEbyTC} ~\citeyear{nick01IEbyTC}. The learning task here
is to classify the lines of text on a business card as a person's name,
affiliation, address, phone number, etc. In this domain, the strong view
consists of the words that appear on each line, based on which a Naive Bayes
text classifier is learned. In the weak view, one can exploit the relative
order of the lines on the card by learning a Hidden Markov Model that
predicts the probability of a particular ordering of the lines on the
business card (e.g., name followed by address, followed by phone number).

This weak view defines a class of concepts that is {\em more general} than
the target concept: all line orderings are possible, even though they are
not equally probable.  Even though the order of the text lines cannot be
used by itself to accurately classify the lines, when combined with the
strong view, the ordering information leads to a classifier that clearly
outperforms the stand-alone strong view \cite{nick01IEbyTC}.

Note that both approaches above use the strong and weak views for passive,
rather than active learning. That is, given a fixed set of labeled and no
unlabeled examples, these algorithms learn one weak and one strong
hypothesis that are then used to craft a {\em domain-specific} predictor
that outperforms each individual hypothesis. In contrast, Co-Testing is an
active learner that seamlessly integrates weak and strong hypotheses without
requiring additional, domain-specific data engineering.

%% file: apps.tex
\section{ Empirical Validation}

In this section we empirically compare Co-Testing with other state of
the art learners. Our goal is to test the following hypothesis: given
a multi-view learning problem, Co-Testing converges faster than its
single-view counterparts.

We begin by presenting the results on three real-world {\em
  classification} domains: Web-page classification, discourse tree
parsings, and advertisement removal. Then we focus on an important
industrial application, {\em wrapper induction}
\cite{stalkerJAAMAS,KushmerickAIJ,SoftMealyJIS98}, in which the goal
is to learn rules that extract the relevant data from a collection of
documents (e.g., extract book titles and prices from a Web site).


The results for classification and wrapper induction are analyzed separately
because:
\begin{description}

\item[-] for each of the three classification tasks, there are only two
strong views that are available; in contrast, for wrapper induction we have
two strong and one weak views, which allows us to explore a wider range of
options.

\item[-] for each classification domain, there is exactly one available
dataset. In contrast, for wrapper induction we use a testbed of 33 distinct
tasks. This imbalance in the number of available datasets requires different
presentation styles for the results.

\item[-] in contrast to typical classification, a major requirement for
wrapper induction is to learn (close to) 100\%-accurate extraction rules
from just a handful of examples \cite[pages 3-6]{MusleaThesis}. This
requirement leads to significant differences in both the experimental setup
and the interpretation of the results (e.g., results that are excellent for
most classification tasks may be unacceptable for wrapper induction).

\end{description}

\begin{table}[thp]
\begin{center}
{\small
\begin{tabular}{|c|c||c|c||c|c|c|c|c|} \hline
            & & \multicolumn{2}{c||}{\bf Co-Testing} &
                \multicolumn{5}{c|} {\bf Single-view Algorithms} \\
                \cline{3-9}
Domain & ${\cal L}$ & Query    & Output    &         &            &         &      &      \\
       &            & Selection& Hypothesis& {\bf QBC} & {\bf qBag} & {\bf qBst} & {\bf US} & {\bf Rnd}\\
\hline
{\sc ad}      &{\sc ib} &  {\em naive}     & {\em winner} & $-$     & $\surd$ & $\surd$ & $\surd$ & $\surd$ \\
\hline
{\sc tf}      &{\sc mc4}&  {\em naive}     & {\em winner} & $-$     & $\surd$ & $\surd$ & $-$ & $\surd$ \\
\hline
              & Naive   &  {\em naive}     &{\em weighted}&         &         &         &         & \\ \cline{3-3}
{\sc courses} & Bayes   &{\em conservative}& {\em vote}   & $\surd$ & $\surd$ & $\surd$ & $\surd$ & $\surd$ \\
\hline
\end{tabular}
}
\caption [Algorithms used on {\sc ad}, {\sc courses}, and {\sc tf}] {The
algorithms used for classification. The last five columns denote
Query-by-Committee/-Bagging/-Boosting, Uncertainty Sampling and Random
Sampling.}
\label{algs}
\end{center}
\end{table}

\subsection{ Co-Testing for Classification}

We begin our empirical study by using three classification tasks to
compare Co-Testing with existing active learners. We first introduce
these three domains and their respective views; then we discuss the
learners used in the evaluation and analyze the experimental results.

\subsubsection{ The Views used by Co-Testing}

We applied Co-Testing to three real-world classification domains for which
there is a natural, intuitive way to create two views:

\begin{description}

\item[-] {\sc ad} \cite{NickAd} is a classification problem with two 
classes, 1500 attributes, and 3279 examples.  In {\sc ad}, images that
appear in Web pages are classified into {\tt ads} and {\tt non-ads}.
The view $V_1$ consists of all textual features that describe the
image; e.g., 1-grams and 2-grams from the caption, from the {\sc url}
of the page that contains the image, from the {\sc url} of the page
the image points to, etc. In turn, $V_2$ describes the properties
of the image itself: length, width, aspect ratio, and ``origin''
(i.e., are the image and the page that contains it coming from the
same Web server?).

\item[-] {\sc courses}~\cite{coTra} is a domain with two classes, 2206
features, and 1042 examples. The learning task consists of classifying
Web pages as {\tt course homepages} and {\tt other pages}. In {\sc
courses} the two views consist of words that appear in the page itself
and words that appear in hyperlinks pointing to them, respectively.

\item[-] {\sc tf}~\cite{tf} is a classification problem with seven
classes, 99 features and 11,193 examples. In the context of a machine
translation system, it uses the shift-reduce parsing paradigm to learn
how to rewrite Japanese discourse trees as English-like discourse
trees.  In this case, $V_1$ uses features specific to the
shift-reduce parser: the elements in the input list and the partial
trees in the stack. $V_2$ consists of features specific to the
Japanese tree given as input.

\end{description}

\subsubsection{ The Algorithms used in the Evaluation}

Table \ref{algs} shows the learners used in this empirical comparison. We
have implemented all active learners as extensions of the {\small ${\cal
MLC}$++} library ~\cite{mlcpp}. For each domain, we choose the base learner
as follows: after applying all primitive learners in ${\cal MLC}${\small ++}
on the dataset (10-fold cross-validation), we select the one that obtains
the best performance. More precisely, we are using the following base
learners: {\sc ib} \cite{ibAha} for {\sc ad}, Naive Bayes \cite{coTra} for
{\sc courses}, and {\sc mc}{\small 4}, which is an implementation of {\sc
c}{\small 4.5}, for {\sc tf}.

The five single-view algorithms from Table \ref{algs} use all
available features (i.e., $V_1 \; \cup \; V_2$) to learn the target
concept.\footnote{ In a preliminary experiment, we have also ran the
algorithms of the individual views. The results on $V_1$ and $V_2$
were either worse then those on $V_1 \; \cup \; V_2$ or the
differences were statistically insignificant. Consequently, for sake
of simplicity, we decided to show here only the single-view results
for $V_1 \; \cup \; V_2$.} On all three domains, Random Sampling ({\bf
Rnd}) is used as strawman; Query-by-Bagging and -Boosting, denoted by
{\bf qBag} and {\bf qBst}, are also run on all three domains. In
contrast, Uncertainty Sampling ({\bf US}) is applied only on {\sc ad}
and {\sc courses} because {\sc mc}{\small 4}, which is the base
learner for {\sc tf}, does not provide an estimate of the confidence
of its prediction.

As there is no known method for {\em randomly} sampling from the {\sc
  ib} or {\sc mc}{\small 4} version spaces, Query-by-Committee ({\bf
  QBC}) is not applied to {\sc ad} and {\sc tf}. However, we apply
{\bf QBC} to {\sc courses} by borrowing an idea from \citeauthor{emAL}
\citeyear{emAL}: we create the committee by sampling hypotheses
according to the (Gamma) distribution of the Naive Bayes parameters
estimated from the training set $L$.

For Query-by-Committee we use a typical 2-hypothesis committee. For
Query-by-Bagging and -Boosting, we use a relatively small 5-hypothesis
committees because of the {\sc cpu} constraints: the running time increases
linearly with the number of learned hypotheses, and, in some domains, it
takes more than 50 {\sc cpu} hours to complete the experiments even with the
5-hypothesis committees.

Because of the limitations of their respective base learners (i.e.,
the above-mentioned issue of estimating the confidence in each
prediction), for {\sc ad} and {\sc tf} we use Naive Co-Testing with a
{\em winner-takes-all} output hypothesis; that is, each query is
randomly selected among the contention points, and the output
hypothesis is the one learned in the view that makes the fewest
mistakes on the queries.  In contrast, for {\sc courses} we follow the
methodology from the original Co-Training paper \cite{coTra}, where
the output hypothesis consists of the {\em weighted vote} of the
classifiers learned in each view.

On {\sc courses} we investigate two of the Co-Testing query selection
strategies: {\em naive} and {\em conservative}. The third, {\em aggressive}
query selection strategy is not appropriate for {\sc courses} because the
``hyperlink view'' is significantly less accurate than the other one (after
all, one rarely encounters more than a handful of words in a hyperlink).
Consequently, most of the high-confidence contention points are ``unfixable
mistakes'' in the hyperlink view, which means that even after seeing the
correct label, they cannot be classified correctly in that view.

\begin{table}[thp]
\begin{center}
\begin{tabular}{|l||c|c|c||c|c|c||c|c|c|} \hline
                & \multicolumn{6}{c||} {\small Naive Co-Testing}
                & \multicolumn{3}{c|} {\small Conservative Co-Testing} \\ \cline{2-10}
{\bf Algorithm} & \multicolumn{3}{c||}{\sc ad} &  \multicolumn{3}{c||}{\sc tf} 
                & \multicolumn{3}{c|} {\sc courses} \\ \cline{2-10}
                & {\small Loss} & {\small Tie} & {\small Win} 
                & {\small Loss} & {\small Tie} & {\small Win} 
                & {\small Loss} & {\small Tie} & {\small Win} \\
\hline
{\small Random Sampling}    
             & 0 &  0 & 19 & 0 & 21 & 70 & 0 &  0 & 49 \\
{\small Uncertainty Sampling}    
             & 0 &  2 & 17 & 0 &  2 & 89 & - &  - &  - \\
{\small Query-by-Committee}    
             & - &  - &  - & 0 & 60 & 31 & - &  - &  - \\
{\small Query-by-Bagging}   
             & 0 & 18 &  1 & 0 &  6 & 85 & 0 & 28 & 21 \\
{\small Query-by-Boosting} 
             & 0 & 15 &  4 & 0 &  0 & 91 & 0 & 0  & 49 \\
\hline
{\small Naive Co-Testing}   
             & - &  - &  - & - &  - &  - & 0 & 21 & 28 \\
\hline
\end{tabular}
\caption[Tests of Statistical Significance]{ Statistical significance
results in the empirical (pair-wise) comparison of the various
algorithms on the three domains.}
\label{statSig}
\end{center}
\end{table}

\subsubsection{ The Experimental Results}

The learners' performance is evaluated based on 10-fold, stratified cross
validation. On {\sc ad}, each algorithm starts with 150 randomly chosen
examples and makes 10 queries after each of the 40 learning episodes, for a
total of 550 labeled examples. On {\sc courses}, the algorithms start with 6
randomly chosen examples and make one query after each of the 175 learning
episodes. Finally, on {\sc tf} the algorithms start with 110 randomly chosen
examples and make 20 queries after each of the 100 learning episodes.

\begin{figure}[thp]
\centerline{\psfig{file=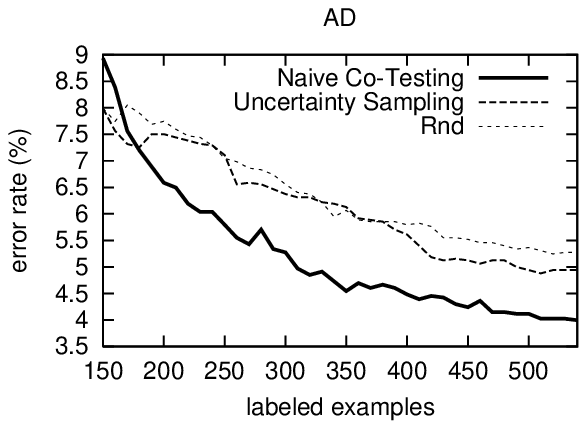,width=10cm}}
\centerline{\psfig{file=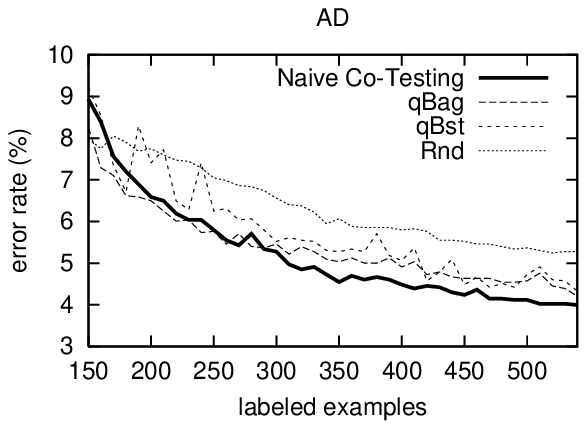,width=10cm}}
\centerline{\psfig{file=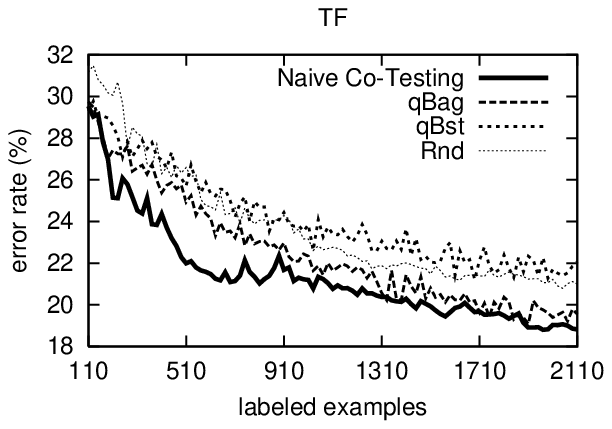,width=10cm}}
\caption{ Empirical results on the {\sc ad} and {\sc tf} problems}
\label{AdTf}
\end{figure}

\begin{figure}[thp]
\centerline{\psfig{file=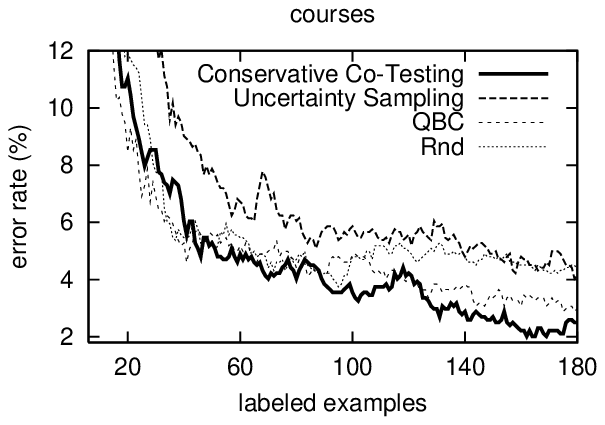,width=10cm}}
\centerline{\psfig{file=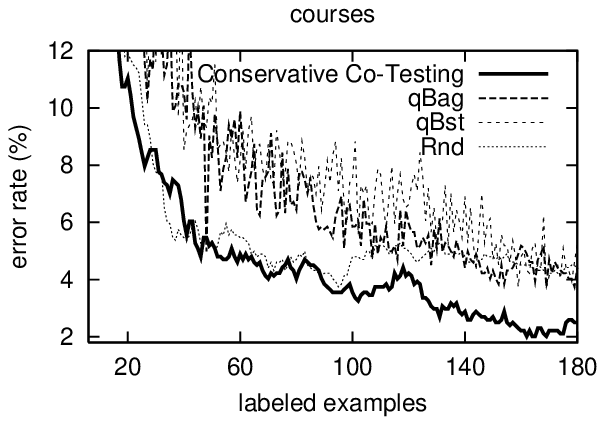,width=10cm}}
\centerline{\psfig{file=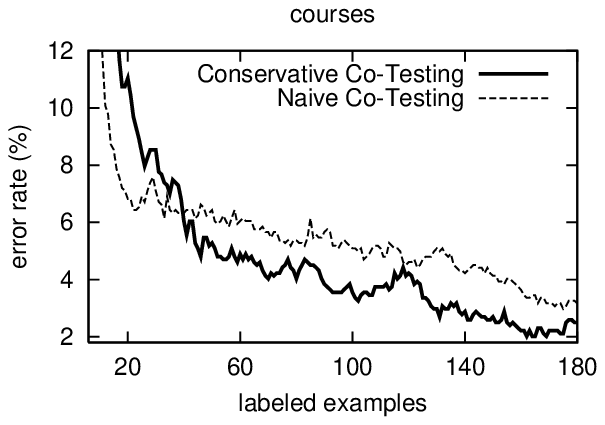,width=10cm}}
\caption{ Empirical results on the {\sc courses} problem}
\label{courses}
\end{figure}

Figures \ref{AdTf} and \ref{courses} display the learning curves of
the various algorithms on {\sc ad}, {\sc tf}, and {\sc course}.  On
all three domains, Co-Testing reaches the highest accuracy (i.e.,
smallest error rate). Table \ref{statSig} summarizes the statistical
significance results ({\tt t}-test confidence of at least 95\%)
obtained in a pair-wise comparison of the various algorithms. These
comparisons are performed on the right-most half of each learning
curve (i.e., towards convergence). The best way to explain the results
in Table ~\ref{statSig} is via examples: the results of comparing
Naive Co-Testing and Random Sampling on {\sc ad} appear in the first
three columns of the first row. The three numbers (i.e., 0, 0, and 19)
mean that on (all) 19 comparison points Naive Co-Testing outperforms
Random Sampling in a statistically significant manner. Similarly,
comparing Naive and Conservative Co-Testing on {\sc courses} (the last
three columns on the last row) leads to the following results: on 28
of the comparison points Conservative Co-Testing outperforms Naive
Co-Testing in a statistically significant manner; on 21 other points
the differences are statistically insignificant; finally, on {\em no}
comparison point Naive Co-Testing outperforms its Conservative
counterpart.

The results in Table \ref{statSig} can be summarized as
follows. First of all, {\em no single-view algorithm} outperforms
Co-Testing in a statistically significant manner on {\em any} of the
comparison points. Furthermore, except for the comparison with
Query-by-Bagging and -Boosting on {\sc ad}, where the difference in
accuracy is statistically insignificant on almost all comparison
points, Co-Testing clearly outperform all algorithms on all domains.

Finally, let us briefly comment on applying multi-view, semi-supervised
learners to the three tasks above. As mentioned in section \ref{coTraRW},
such algorithms bootstrap the views from each other by training each view on
the examples labeled with high-confidence by the other view. For {\sc ad}
and {\sc tf}, we could not use multi-view, semi-supervised learning because
the base learners {\sc ib} and {\sc mc}{\small 4} do not provide a
(reliable) estimate of the confidence in their predictions. More precisely,
{\sc mc}{\small 4} provides no estimate at all, while {\sc ib}'s estimates
are extremely poor when the training data is scarce (e.g., see the poor
performance of Uncertainty Sampling on {\sc ad}, where it barely outperforms
Random Sampling).

On {\sc courses}, we have applied both Co-Training and Co-EM in
conjunction with the Naive Bayes base learner. Both these multi-view
learners reach their maximum accuracy (close to 95\%) based on solely
12 labeled and 933 unlabeled examples, after which their performance
does not improve in a statistically significant manner.\footnote{ A
recent paper \cite{coEMSVM} shows that - for text classification -
{\sc svm} is more appropriate than Naive Bayes as base learner for
Co-EM, though not necessarily for Co-Training. As the ${\cal MLC}${\small ++} 
library does not provide {\sc svm} as base learner,
we could not compare our results with those by \citeauthor{coEMSVM}
\citeyear{coEMSVM}, where Co-EM + {\sc svm} reaches 99\% accuracy
based on 12 labeled and 933 unlabeled examples. However, in all
fairness, it is unlikely that Co-Testing could lead to an even faster
convergence.}  As shown in Figure \ref{courses}, when the training
data is scarce (i.e., under 40 labeled examples), Co-Testing's
accuracy is less than 95\%; however, after making additional queries,
Co-Testing reaches 98\% accuracy, while Co-Training and Co-EM remain
at 95\% even when trained on 180 labeled and 765 unlabeled examples.
These results are consistent with the different goals of active and
semi-supervised learning: the former focuses on learning the {\em
  perfect} target concept from a minimal amount of labeled data, while
the latter uses unlabeled examples to boost the accuracy of a
hypothesis learned from just a handful of labeled examples.

\subsection{ Co-Testing for Wrapper Induction}  
\label{experimWI}

We focus now on a different type of learning application, {\em wrapper
induction} ~\cite{stalkerJAAMAS,KushmerickAIJ}, in which the goal is
to learn rules that extract relevant sub-strings from a collection of
documents. Wrapper induction is a key component of commercial systems
that integrate data from a variety of Web-based information sources.

\subsubsection{ The Views used by Co-Testing}

\begin{figure}[t]
\centerline{ \psfig{file=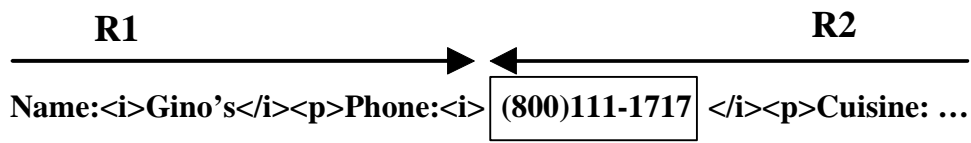} }
\caption[Forward and backward extraction rules] 
{ Both the {\em forward} and {\em backward} rules detect the beginning of
the phone number.}
\label{intuitFBRules}
\end{figure}

Consider the illustrative task of extracting phone numbers from
documents similar to the fragment in Figure ~\ref{intuitFBRules}. To
find where the phone number begins,\footnote{ As shown by
~\citeauthor{stalkerJAAMAS} ~\citeyear{stalkerJAAMAS}, the end of the
phone number can be found in a similar manner.} one can use the rule

\begin{description}
\item[] {\bf R1} = $SkipTo${\bf (} {\small {\tt Phone:}\verb+<+{\tt i}\verb+>+} {\bf )}
\end{description}

\noindent This rule is applied {\em forward}, from the beginning of the
page, and it ignores everything until it finds the string {\tt
Phone:}\verb+<+{\tt i}\verb+>+. Note that such forward-going rules do not
represent the only way to detect where the phone number begins: an
alternative approach is to use the rule
 
\begin{description}
\item[] {\bf R2} = $BackTo${\bf (} {\small {\tt Cuisine}} {\bf )}
                   $BackTo${\bf (} {\small ( {\em Number} )} {\bf )}
\end{description}

\noindent which is applied {\em backward}, from the $end$ of the document.
{\bf R2} ignores everything until it finds ``{\tt Cuisine}'' and then,
again, skips to the first number between parentheses.

Forward and backward rules such as {\bf R1} and {\bf R2} can be
learned from user-provided examples by the state of the art wrapper
induction system {\sc stalker} \cite{stalkerJAAMAS}, which we use as
base learner for Co-Testing. Intuitively, {\sc stalker} creates a
forward or a backward rule that consumes all the tokens that precede
or follow the extraction point, respectively. It follows that rules
such as {\bf R1} and {\bf R2} represent descriptions of the same
concept (i.e., beginning of phone number) that are learned in two
different views: the sequences of tokens that {\em precede} and {\em
  follow} the beginning of the item, respectively. These views are
{\em strong views} because each of them is sufficient to accurately
extract the items of interest ~\cite{stalkerJAAMAS,coTst}.

In addition to these two views, which rely mostly on the {\em context} of
the item to be extracted (i.e., the text surrounding the item), one can use
a third view that describes the content of the item to be extracted. For
example, phone numbers can be described by the simple grammar: ``{\tt (}
{\em Number} {\tt )} {\em Number} {\tt -} {\em Number}''; similarly, most
{\sc url}s start with ``{\tt http://www.}'', end with ``{\tt .html}'', and
contain no {\sc html} tags.

Such a content-based view is a {\em weak view} because it often represents a
concept {\em more general} than the target one. For example, the phone
number grammar above cannot discriminate between the home, office, cell, and
fax numbers that appear within the same Web page; similarly, the {\sc url}
grammar cannot distinguish between the {\sc url}s of interest (e.g., a
product's review) and all the other ones (e.g., advertisements).

In this weak view, we use as base learner a version of DataPro
\cite{lerman2003} that is described elsewhere \cite{muslea2003}.
DataPro learns - from positives examples only - the ``prototypes'' of
the items to be extracted; i.e., it finds statistically significant
sequences of tokens that (1) are highly unlikely to have been
generated by chance and (2) describe the content of many of the
positive examples.  The features used by this base learner consist of
the {\em length range} (in tokens) of the seen examples, the {\em
  token types} that appear in the training set (e.g., {\em Number},
{\em AllCaps}, etc), and the {\em start-} and {\em end-} pattern
(e.g., ``{\tt http://www.}''  and ``{\em AlphaNum} {\tt .html}'',
respectively).

\subsubsection{ The Algorithms used in the Evaluation}
\label{AL4WI}

As the extraction rules learned in the two strong views do not provide any
estimate of the confidence of the extractions, the only Co-Testing algorithm
that can be implemented based solely on the forward and backward views is
{\em Naive Co-Testing} with a {\em winner-takes-all} output hypothesis:
\begin{description}
\item[-] each query is randomly chosen ({\em Naive Co-Testing}) among the
contention points, which are the documents from which the learned rules
extract different strings.

\item[-] the output hypothesis is the rule learned in the view that makes
the fewest mistakes over the allowed number of queries (i.e., {\em
winner-takes-all}).
\end{description}

Given the additional, content-based view, we can also implement an {\em
aggressive} version of Co-Testing for wrapper induction:
\begin{description}

\item[-] the contention points are, again, the unlabeled examples on which
the rules learned in the {\em strong} views do not extract the same string.
 
\item[-] the {\em aggressive query selection strategy} works by
selecting the contention point for which the hypothesis learned in the
weak view is maximally confident that {\em both} {\sc stalker} rules
are extracting incorrect strings. More formally, for each contention
point, let $s_1$ and $s_2$ be the strings extracted by the strong
views; let us also denote by $n_1$ and $n_2$ the number of constraints
learned in the weak views that are violated by $s_1$ and $s_2$. Using
this notation, the next query is the contention point for which
$min(n_1,n_2)$ has the largest value.

\item[-] the output hypothesis is obtained by the following {\em majority
voting} scheme: if the strings extracted by the strong views are identical,
then they represent the extracted item; otherwise the result is the one of
these two strings that violates fewer of the constraints learned in the weak
view.
\end{description}

In the empirical evaluation below, we compare these two Co-Testing
algorithms with Random Sampling and Query-by-Bagging. The former is used as
strawman, while the latter is the only general-purpose active learner that
can be applied in a straightforward manner to wrapper induction (for
details, see the discussion in section \ref{coTstVsAL}). Finally, existing
multi-view, semi-supervised learners cannot be used for wrapper induction
because the base learners do not provide an estimate in the confidence of
each extraction; and even if such an estimate could be obtained, wrapper
induction algorithms are extremely sensitive to mislabeled examples, which
would make the bootstrapping process unacceptably brittle.

In this paper, the implementation of Random Sampling is identical with
that of Naive Co-Testing with {\em winner takes all}, except that it
randomly queries one of the unlabeled examples from the working set.
For Query-by-Bagging, the committee of hypotheses is created by
repeatedly re-sampling (with substitution) the examples in the
original training set $L$. We use a relatively small committee (i.e.,
10 extraction rules) because when learning from a handful of examples,
re-sampling with replacement leads to just a few distinct training
sets.  In order to make a fair comparison with Co-Testing, we run
Query-by-Bagging once in each strong view and report the best of the
obtained results.

\subsubsection{ The Experimental Results}

\begin{table}[thp]
\begin{center}
\small{
\centerline{
{
\begin{tabular}{|c|c|c||c||} \hline
 Task & Source & Item & Nmb \\
   ID & name   & name & exs \\       
\hline
\hline
 S1-0 &                 & Price        & 404 \\
 S1-1 & Computer        & URL          & 404 \\
 S1-2 & ESP             & Item         & 404 \\
\hline
 S2-0 &                 & URL          & 501 \\
 S2-1 & CNN/Time        & Source       & 501 \\
 S2-2 & AllPolitics     & Title        & 499 \\
 S2-3 &                 & Date         & 492 \\
\hline
 S3-0 &                 & URL          & 175 \\
 S3-1 &                 & Name         & 175 \\
 S3-3 & Film.com        & Size         & 175 \\
 S3-4 & Search          & Date         & 175 \\
 S3-5 &                 & Time         & 175 \\
\hline
 S6-1 & PharmaWeb       & University   &  27 \\
\hline
S9-10 & Internet        & Arrival Time &  44 \\
S9-11 & Travel Net.     & Availability &  39 \\
\hline
S11-1 & Internet        & Email        &  91 \\ 
S11-2 & Address         & Update       &  91 \\ 
\hline
\end{tabular}
}
~
{
\begin{tabular}{|c|c|c||c||} \hline
 Task & Source & Item & Nmb \\
   ID & name   & name & exs \\       
\hline
\hline
S11-3 & Finder          & Organization &  72 \\
\hline
S15-1 & NewJour         & Name         & 355 \\  
\hline
S19-1 & Shops.Net       & Score        & 201 \\
S19-3 &                 & Item Name    & 201 \\
\hline
S20-3 & Democratic      & Score        &  91 \\
S20-5 & Party Online    & File Type    & 328 \\
\hline
S24-0 & Foreign         & Language     & 690 \\ 
S24-1 & Languages for   & URL          & 424 \\ 
S24-3 & Travelers       & Translation  & 690 \\
\hline
S25-0 & {\sc us} Tax Code & URL          & 328 \\
\hline
S26-3 &                   & Price        & 377 \\
S26-4 & CD Club           & Artist       & 377 \\  
S26-5 & Web Server        & Album        & 377 \\
\hline
S28-0 & Cyberider         & URL          & 751 \\
S28-1 & Cycling {\sc www} & Relevance    & 751 \\
\hline
S30-1 & Congress          & Person Name  &  30 \\
      & Quarterly         &              &     \\
\hline
\end{tabular}
}
}
}
\caption{ The 33 wrapper induction tasks used in the empirical evaluation.}
\label{new33tasks}
\end{center}
\end{table}

In our empirical comparison, we use the 33 most difficult wrapper
induction tasks from the testbed introduced by Kushmerick (1998,
2000). These tasks, which were previously used in the literature
\cite{muslea2003,MusleaThesis}, are briefly described in Table
\ref{new33tasks}. We use 20-fold cross-validation to compare the
performance of Naive and Aggressive Co-Testing, Random Sampling, and
Query-by-Bagging on the 33 tasks. Each algorithm starts with two
randomly chosen examples and then makes 18 successive queries.

The results below can be summarized as follows: for 12 tasks, only the two
Co-Testing algorithms learn 100\% accurate rules; for another 18 tasks,
Co-Testing and {\em at least} another algorithm reach 100\% accuracy, but
Co-Testing requires the smallest number of queries. Finally, on the
remaining three tasks, no algorithm learns a 100\% accurate rule.

\begin{figure}[thp]
\centerline{\psfig{file=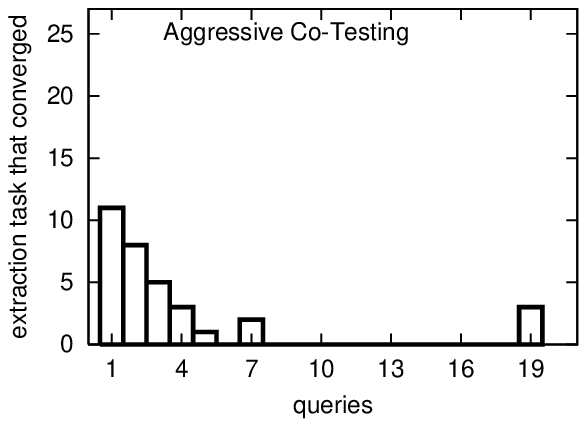,width=7cm} ~
	    \psfig{file=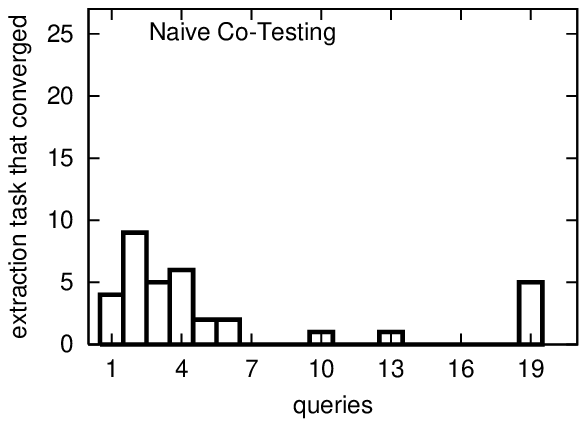,width=7cm} }
	    \centerline{\psfig{file=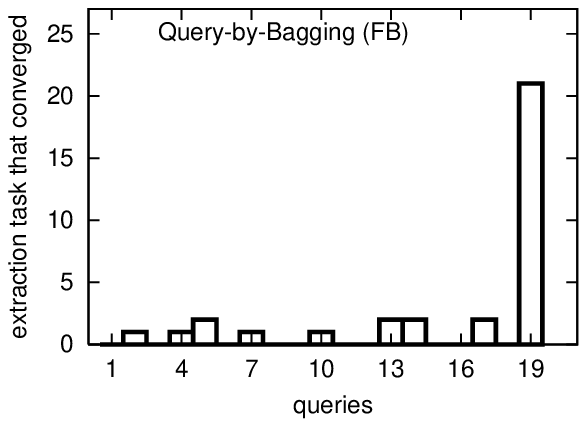,width=7cm} ~
	    \psfig{file=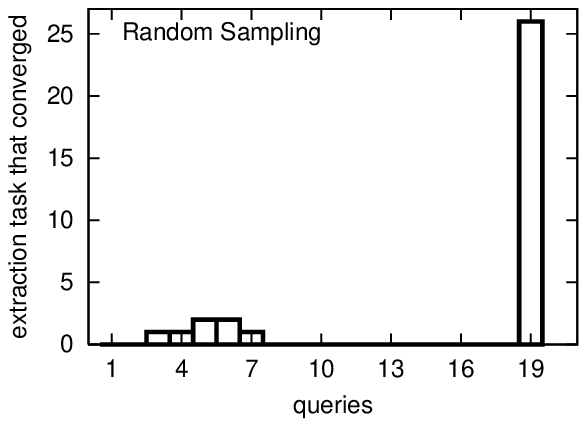,width=7cm} }
\caption[Convergence results for the 33 wrapper induction tasks]{ 
Convergence results for the 33 wrapper induction tasks.}
\label{convergence}
\end{figure}

Figure \ref{convergence} shows the aggregate performance of the four
algorithms over the 33 tasks. In each of the six graphs, the {\tt X} axis
shows the number of queries made by the algorithm, while the {\tt Y} axis
shows the number of tasks for which a 100\% accurate rule was learned based
on exactly {\tt X} queries. As mentioned earlier, all algorithms start with
2 random examples and make 18 additional queries, for a total of 20 labeled
examples. {\em By convention}, the right-most point on the {\tt X} axis,
which is labeled ``{\small {\bf {\tt 19 queries}}}'', represents the number
of tasks that require more than the allowed 18 queries to learn a 100\%
accurate rule. This additional ``{\small {\bf {\tt 19 queries}}}''
data-point allows us to summarize the results without dramatically extending
the {\tt X} axis beyond 18 queries: as for some of the extraction tasks
Random Sampling and Query-by-Bagging need hundreds of queries to learn the
correct rules, the histograms would become difficult to read if the entire
{\tt X} axis were shown.

As shown in Figure \ref{convergence}, the two Co-Testing algorithms clearly
outperform their single-view counterparts, with Aggressive Co-Testing doing
significantly better than Naive Co-Testing (the results are statistically
significant with a confidence of at least 99\%).  Aggressive Co-Testing
learns 100\%-accurate rules on 30 of the 33 tasks; for all these tasks, the
extraction rules are learned from at most seven queries.  Naive Co-Testing
learns 100\% accurate rules on 28 of the 33 tasks.  On 26 of these 28 tasks,
the extraction rules are learned based on at most six queries.  In contrast,
Random Sampling and Query-by-Bagging learn 100\% accurate rules for only
seven and twelve of the tasks, respectively.  In other words, both
Co-Testing algorithms learn the correct target concept for more than twice
as many tasks than Query-by-Bagging or Random Sampling.

We must emphasize the power of Aggressive Co-Testing on high-accuracy tasks
such as wrapper induction: for 11 of the 33 tasks, a single,
``aggressively-chosen'' query is sufficient to learn the correct extraction
rule. In contrast, Naive Co-Testing converges in a single query on just four
of the 33 tasks, while the other two learners never converge in a single
query.

For the three tasks on which Aggressive Co-Testing does not learn 100\%
accurate rules, the failure is due to the fact that one of the views is
significantly less accurate than the other one. This leads to a majority of
contention points that are mislabeled by the ``bad view,'' which - in turn -
skews the distribution of the queries towards mistakes of the ``bad view.''
Consequently, Co-Testing's performance suffers because such queries are
uninformative for both views: the ``good view'' makes the correct prediction
on them, while the ``bad view'' is inadequate to learn the target
concept. In order to cope with this problem, we introduced a view validation
algorithm \cite{vValidICML02} that predicts whether the views are
appropriate for a particular task.

Finally, let us briefly compare the results above with the ones obtained by
{\sc wien} \cite{KushmerickAIJ}, which is the only wrapper induction system
that was evaluated on all the extraction tasks used here. As the two
experimental setups are not identical (i.e., cross-validation vs. random
splits) this is just an {\em informal} comparison; however, it puts our
results into perspective by contrasting Co-Testing with another state of the
art approach to wrapper induction.

The results can be summarized as follows: {\sc wien} fails on 18 of the 33
task; these 18 tasks include the three for which Aggressive and Naive
Co-Testing failed to learn perfect rules.  On the remaining 15 tasks, {\sc
wien} requires between 25 and 90 examples\footnote{ In the {\sc wien}
framework, an example consists of a document in which {\em all} items of
interest are labeled. For example, a page that contains a list of 100 names,
all labeled, represents a single labeled example. In contrast, for {\sc
stalker} the same labeled document represents 100 distinct labeled
examples. In order to compare the {\sc wien} and {\sc stalker} results, we
convert the {\sc wien} data to {\sc stalker}-like data by multiplying the
number of labeled {\sc wien} pages by the average number of item occurrences
in each page.} to learn the correct rule. For the same 15 tasks, both
Aggressive and Naive Co-Testing learn 100\% accurate rules based on at most
eight examples (two random plus at most six queries).

%% file: Conclusions.tex
\section{ Conclusion}

In this paper we introduce Co-Testing, which is an active learning
technique for multi-view learning tasks. This novel approach to active
learning is based on the idea of {\em learning from mistakes}; i.e.,
Co-Testing queries unlabeled examples on which the views predict a
different label (such contention points are guaranteed to represent
mistakes made in one of the views). We have analyzed several members
of the Co-Testing family (e.g., Naive, Conservative and Aggressive
Co-Testing). We have also introduced and evaluated a Co-Testing
algorithm that simultaneously exploits both strong and weak views.


Our empirical results show that Co-Testing is a powerful approach to active
learning. Our experiments use four extremely different base learners (i.e.,
{\sc stalker}, {\sc ib}, Naive Bayes, and {\sc mc}{\small 4}) on four
different types of domains: wrapper induction, text classification ({\sc
courses}), ad removal ({\sc ad}), and discourse tree parsing ({\sc tf}). In
all these scenarios, Co-Testing clearly outperforms the single-view, state
of the art active learning algorithms. Furthermore, except for
Query-by-Bagging, Co-Testing is the only algorithm that can be applied to
all the problems considered in the empirical evaluation. In contrast to
Query-by-Bagging, which has a poor performance on {\sc courses} and wrapper
induction, Co-Testing obtains the highest accuracy among the considered
algorithms.

Co-Testing's success is due to its ability to discover the mistakes made in
each view. As each contention point represents a mistake (i.e., an erroneous
prediction) in at least one of the views, it follows that each query is
extremely informative for the view that misclassified that example; that is,
mistakes are more informative than correctly labeled examples. This is
particularly true for base learners such as {\sc stalker}, 
which do not improve the current hypothesis unless they are provided with
examples of misclassified instances.

As a limitation, Co-Testing can be applied only to multi-view tasks; that
is, unless the user can provide two views, Co-Testing cannot be used at
all. However, researchers have shown that besides the four problems above,
multiple views exist in a variety of real world problems, such as named
entity classification \cite{coBoost}, statistical parsing
\cite{coTraSarkar}, speech recognition \cite{speech-2V}, word sense
disambiguation \cite{yarowsky-2v}, or base noun phrase bracketing
\cite{pierce01Limit}.

The other concern about Co-Testing is related to the potential violations of
the two multi-view assumptions, which require that the views are both
uncorrelated and compatible. For example, in the case of correlated views,
the hypotheses learned in each view may be so similar that there are no
contention points among which to select the next query. In terms of view
incompatibility, remember that, for three of the 33 wrapper induction tasks,
one of the views was so inaccurate that the Co-Testing could not outperform
Random Sampling. In two companion papers \shortcite{coEMT,vValidICML02} we
have proposed practical solutions for both these problems.